# Mechatronic Design, Experimental Setup and Control Architecture Design of a Novel 4 DoF Parallel Manipulator

Marina Vallés[a], Pedro Araujo-Gómez[b], Vicente Mata[c], Angel Valera[a*], Miguel Díaz-Rodríguez[b], Álvaro Page[d] and Nidal M. Farhat[e]

[a] *Instituto Universitario de Automática e Informática Industrial, Universitat Politècnica de València, Camino de Vera s/n, 46022 Valencia, Spain. E-mail: {mvalles, giuprog}@isa.upv.es*

[b] *Laboratorio de Mecatrónica y Robótica, Facultad de Ingeniería, Universidad de los Andes, La Hechicera, 5101 Mérida, Venezuela. E-mail: {pfaraujo, dmiguel}@ula.ve*

[c] *Centro de Investigación en Ingeniería Mecánica. Universitat Politècnica de València, Camino de Vera s/n, 46022 Valencia, Spain. E-mail: vmata@mcm.upv.es*

[d] *Instituto de Biomecánica de Valencia. Universitat Politècnica de València, Camino de Vera s/n, 46022 Valencia, Spain. E-mail: alvaro.page@ibv.upv.es*

[e] *Faculty of Engineering & Technology, Dpt. of Mechanical and Mechatronics Engineering, Birzeit, Palestine. E-mail: nfarhat@birzeit.edu*

*corresponding author

# Mechatronic Design, Experimental Setup and Control Architecture Design of a Novel 4 DoF Parallel Manipulator

Although parallel manipulators (PMs) started with the introduction of architectures with 6 Degrees of Freedom (DoF), a vast number of applications require less than 6 DoF. Consequently, scholars have proposed architectures with 3 DoF and 4 DoF, but relatively few 4 DoF PMs have become prototypes, especially of the two rotation (2R) and two translation (2T) motion types. In this paper, we explain the mechatronics design, prototype and control architecture design of a 4 DoF PM with 2R2T motions. We chose to design a 4 DoF manipulator based on the motion needed to complete the tasks of lower limb rehabilitation. To the author's best knowledge, PMs between 3 and 6 DoF for rehabilitation of lower limb have not been proposed to date. The developed architecture enhances the three minimum DoF required by adding a 4 DoF which allows combinations of normal or tangential efforts in the joints, or torque acting on the knee. We put forward the inverse and forward displacement equations, describe the prototype, perform the experimental setup, and develop the hardware and control architecture. The tracking accuracy experiments from the proposed controller show that the manipulator can accomplish the required application.

**Introduction**

From academia to industry, Parallel Manipulators (PMs) have received a great deal of attention and have become a very active area of research. Examples of PM-based applications can be found as flight and motion simulations (Tsai, 1999), food manipulators (Xu et al., 2008), medical applications (Li and Xu, 2007), milling machines (Pierrot and Company, 1999), assembly manipulators (Chablat and Wenger, 2003), robotic rehabilitation (Vallés et al., 2015), among others.

In terms of the PM architecture, the first of its kind consisted of a based platform connected through six (6) limbs to a mobile platform. The legs arrangement provided 6 Degrees of Freedom (DoF) to the end-effector located on the mobile platform (Gough and Whitehall, 1962) and (Stewart, 1965). This architecture is still applied today to develop new applications, and thus new strategies for designing PM is a topic of continuous research (Cao et al., 2015). However, since many applications require less than 6 DoF, new architectures with less DoF called limited DOF PM have been developed. One advantage of designing limited DoF PM is that they maintain some advantages of 6 DoF while reducing development-cost (designing, manufacturing and operation). Examples of this kind of PM are the Delta Robot with three translational DoF (3T) (Clavel, 1988), or the 3-RPS (Lee and Arjunan, 1992), (Carretero et al., 2000). There is also the 3-PRS with two rotational motions and one translational DoF (2R1T) (Chablat and Wenger, 2003), (Vallés et al., 2012), where R, P and S stand for the revolute, prismatic and spherical joints, respectively. Some scholars have proposed a subset of platforms with 4 DoF, mainly for flight simulation purpose, and with three rotational and one translation DOF (3R1T parallel manipulators). Nevertheless, the literature regarding 4 DoF PMs is limited compared with the series of 6, 3 and 2 DoF (Zarkandi, S. 2011). More recently, (Gan et al. 2015) proposed a 2RPS-2UPS architecture to deal with automating fiber placement for aerospace part manufacturing. Among the 4 DoF (without actuation redundancy), we found in the literature that very few of them have become actual prototypes, and in the field of rehabilitation we found that a reconfigurable manipulator with 4 DoF was built (Yoon et al., 2006).

Nowadays, PM are emerging as a conceptual design in the field of rehabilitation robotics (Cazalilla et al., 2016). In the field of lower limb rehabilitation (LLR), most of the PMs developed to date consist of 2 and 3 rotational DoF, mainly because they focus on ankle rehabilitation (Jamwal et al., 2015). Girone et al., (2001) proposed a 6 DoF as a lower limb rehabilitation, although the authors basically adapted a Gough PM architecture for the required task. The above architectures can be suitable for very restricted motions such as the one which takes place in ankle rehabilitation. However, they cannot be extended to rehabilitation of other joints such as the knee or hip. These joints require flexion-extension motion in the sagittal plane, as well as small rotations involving systems with three or more degrees of freedom, of which at least two must be translational motion. A 6 DOF PM can be seen as a first design concept for LLR (Rastegarpanah et al., 2016). As we mentioned before, this solution increases cost and requires an intricate control and dynamic robot model (Janmwal et al, 2015). We are interested in developing a relatively simpler solution.

In order to look for a simpler solution, we need to establish the essential motion which takes place in the LLR. In this regard, the task requires at least 3 DoF, i.e. 2 translations for planar motion and one rotation for flexion-extension motion (Araujo-Gómez et al., 2016), (Mohan et. al, 2017). To the authors' best knowledge, we have not found PMs for LLR between 3 and 6 DoF. We have found serial manipulators which are exoskeleton-based, allowing motions that are compatible with lower limb joint motions (Díaz et al., 2011). Conversely, exoskeleton is unable to deal with combinations of normal or tangential efforts in the joint, or torque acting on the knee, which limits the ability to portray some of the rehabilitation and diagnosis tasks for the knee joint. For instance, wall squats, decline eccentric squats, exercises that involve applying a relevant force in the anteroposterior, or the ability to control the torque applied to the knee (Escamilla et al., 2012).

In this paper, we present the mechatronic design of a two translational (2T) and two rotational (2R) 4 DoF PM which is able to carry out a large number of procedures applicable to LLR, where the mobile platform can simulate the foot trajectory during physiotherapy exercises. We also present the experimental setup including the control architecture design. The main contributions of our paper are the following: 1) the developed architecture enhances the three minimum DoF required by adding a 4 DoF which allows combinations of normal or tangential efforts in the joint, or torque acting on the knee. 2) The robot is able to apply torque to the ligaments of the knee joint without parasite motion on the end-effector. 3) Although many published papers deal with 4 DoF and present the kinematics and dynamics analysis, few prototypes have been built and few have provided its experimental setup.

**Parallel Manipulator Design**

*Presentation of the 4 DoF parallel manipulator*

We have taken the following guidelines into account when designing the manipulator:

- The manipulator should bear a ratio of the person's weight. In addition, the device should be portable and its size as small as possible. As a design concept, a PM meets the specification.

- One of the legs of the PM should be located in the centre of the mobile platform to bring both stability and load capacity to the manipulator.

- The end-effector should be able to move with planar motion on the plane defined by the axis which is normal to the sagittal plane. In addition, it should have two rotations, one parallel to the y-axis and the second one which is normal to the moving platform. A RPU central leg constrains the end-effector to move in a plane (sagittal plane), the U joint defines the rotational DoF.

- The 4 DoF can be achieved with 4 legs (Merlet, J.P., 2006). Therefore, the manipulator should have three additional legs. Since the central leg constrains the end-effector to the required DoF, the external legs should allow 6 DoF. A UPS leg is considered.

- The spherical (S) and the universal (U) joints located on the mobile platform should lie in the same plane, thus avoiding or reducing parasite motions on the end-effector.

Figure 1 shows the 3UPS+RPU which consists of four legs equipped with an active prismatic joint (P). Figure 2 shows the actual PM and its schematic representation. The legs are located as follows: three identical 3-UPS external limbs (U stands for Universal joint, the underlying letter P indicates the actuated joint), and a central RPU limb. The external limbs are equally spaced around the central limb at a radius $r$ in the case of the fixed base and a radius $r_m$ for the mobile platform (see Figure 2).

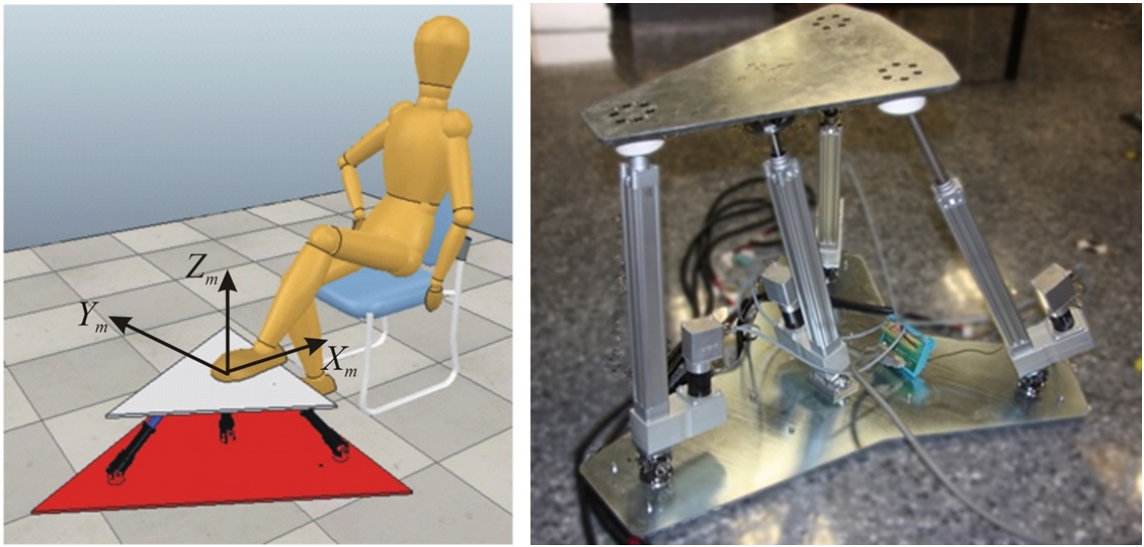

Figure 1. Virtual and actual 4 DoF parallel manipulator.

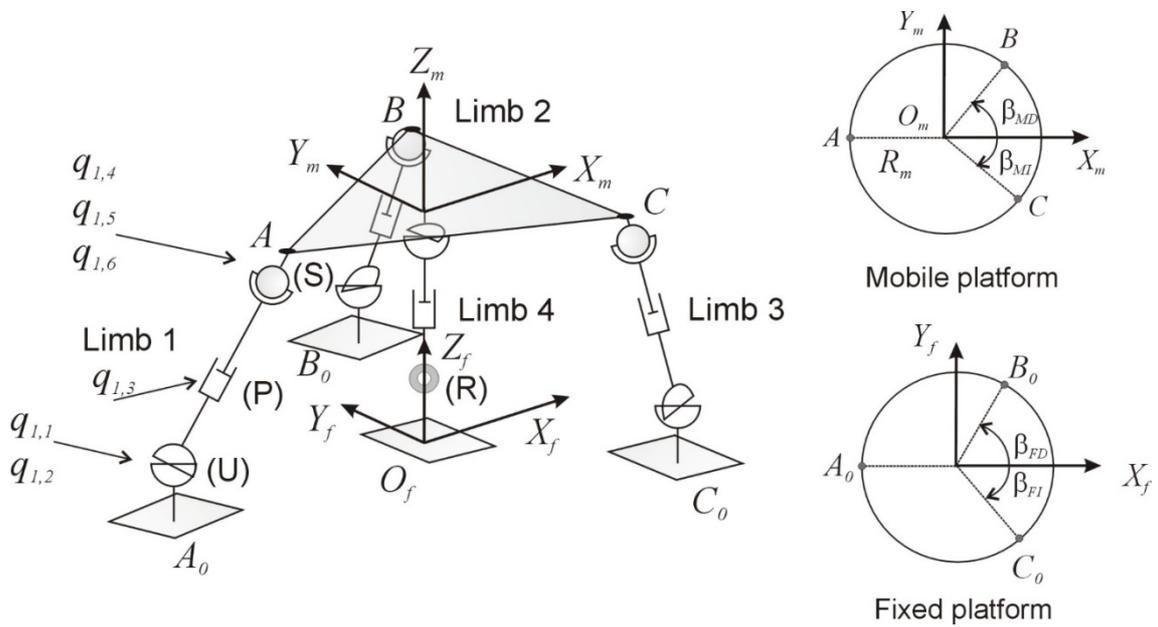

Figure 2. Actual PM and localization of the coordinate systems.

Figure 2 also shows the arrangement of the kinematic pairs. The first axis of rotation of the U-joints, located at the base, points parallel to the axis of the central R joint. In the same figure, the reference systems attached both to the fixed base and to the mobile platform are also depicted. The proposed PM is made up of nine mobile links, five type I kinematic joints (four prismatic + one rotational), four type II kinematic joints (universal) and three type III kinematic joints (spherical). Following the Grübler Kutzbach criterion, the PM has four degrees of freedom.

Figure 3 shows the actual PM in more detail. Each external limb consists of: A) a universal joint connecting the fixed platform to the limb, B) a prismatic joint actuated by the DC motor, and C) a spherical passive joint connecting the limb to the mobile platform. The central limb consists of: D) a passive revolute joint connecting the fixed platform to the limb, E) a prismatic joint controlled by the DC motor, and F) a universal passive joint connecting the limb to the mobile platform.

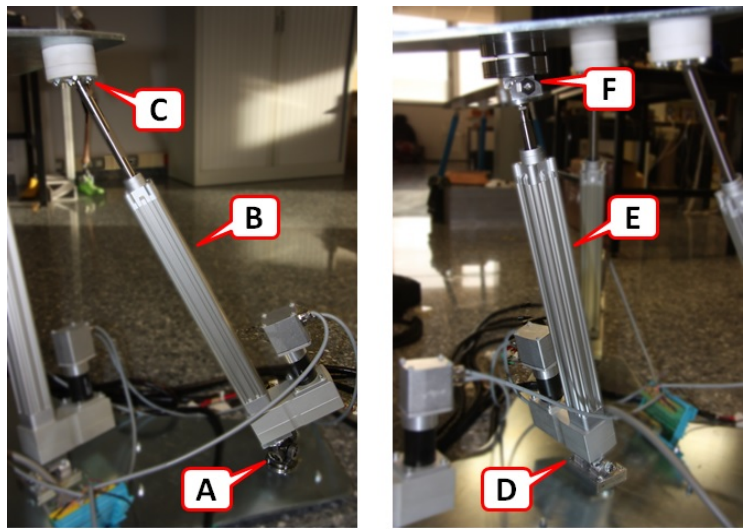

Figure 3. Detailed parts of the actual PM.

Table 1 shows the D-H parameters for the external legs of the actual PM. The subscript *i,j* denotes the joint j on limb *i*. Figure 2 shows the parameters corresponding to leg 1. Table 2 shows the D-H parameters for the central leg. In both cases, we use Paul's notation (Paul, 1981).

**Table 1** D-H Parameter for the UPS limbs (for $i = 1..3$) of the 4 DoF PM

| j | $\alpha_{ij}$ | $a_{ij}$ | $d_{ij}$ | $\theta_{ij}$ |
|---|---|---|---|---|
| 1 | $-\pi/2$ | 0 | 0 | $q_{i,1}$ |
| 2 | $\pi/2$ | 0 | 0 | $q_{i,2}$ |
| 3 | 0 | 0 | $q_{i,3}$ | 0 |
| 4 | $\pi/2$ | 0 | 0 | $q_{i,4}$ |
| 5 | $\pi/2$ | 0 | 0 | $q_{i,5}$ |
| 6 | $\pi/2$ | 0 | 0 | $q_{i,6}$ |

**Table 2** D-H Parameter for the RPS limb of the 4 DoF PM

| j | $\alpha_{ij}$ | $a_{ij}$ | $d_{ij}$ | $\theta_{ij}$ |
|---|---|---|---|---|
| 1 | $-\pi/2$ | 0 | 0 | $q_{4,1}$ |
| 2 | $\pi/2$ | 0 | $q_{4,2}$ | $\pi$ |
| 3 | $\pi/2$ | 0 | 0 | $q_{4,3}$ |
| 4 | 0 | 0 | 0 | $q_{4,4}$ |

*4 DoF parallel manipulator inverse kinematics*

Given the rotational (pitch ($\beta$) and yaw ($\psi$)) angles, and the translations in the $X_f$-$Z_f$ plane, the inverse position equations consist of finding the linear displacement of the actuators: $q_{i,3}$; $i = 1..3$; for the external limbs (UPS) and $q_{4,2}$ for the central limb (RPU). This problem will be divided into two parts: first, we obtain the UPS limb coordinates $q_{i,1}$, $q_{i,2}$, $q_{i,3}$, $i = 1..3$, and the central RPU limb coordinates $q_{4,1}$, $q_{4,2}$. Secondly, we obtain the passive coordinates ($q_{i,4}, q_{i,5}, q_{i,6}$, $i = 1..3$) of the UPS limbs and $q_{4,3}$ and $q_{4,4}$ of RPU.

In order to define the orientation and translation of frame *j* with regard to the *j-1* for the *i-th* limb, the following transformation matrix can be used:

$$^{j-1}H_j^i = \begin{bmatrix} C_{\theta_{ij}} & -C_{\alpha_{ij}} \cdot S_{\theta_{ij}} & S_{\alpha_{ij}} \cdot S_{\theta_{ij}} & a_{ij} \cdot C_{\theta_{ij}} \\ S_{\theta_{ij}} & C_{\alpha_{ij}} \cdot C_{\theta_{ij}} & -S_{\alpha_{ij}} \cdot C_{\theta_{ij}} & a_{ij} \cdot S_{\theta_{ij}} \\ 0 & S_{\alpha_{ij}} & C_{\alpha_{ij}} & d_{ij} \\ 0 & 0 & 0 & 1 \end{bmatrix} \quad (1)$$

where *S* and *C* stand for sine and cosine of the corresponding angle. The closure equation for the central limb can be written as follows:

$$\begin{bmatrix} x_m \\ y_m \\ z_m \end{bmatrix} = \left[ {}^f H_0^4 \cdot {}^0 H_1^4(q_{4,1}) \cdot {}^1 H_2^4(q_{4,2}) \right]_{4,2}^{4,1} \quad (2)$$

for the other limbs, the closure equations can be established as follows,

$$\begin{bmatrix} x_m \\ y_m \\ z_m \end{bmatrix} + {}^f R_m \cdot \begin{bmatrix} -r_m \\ 0 \\ 0 \end{bmatrix} = \left[ {}^f H_0^1 \cdot {}^0 H_1^1(q_{1,1}) \cdot {}^1 H_2^1(q_{1,2}) \cdot {}^2 H_3^1(q_{1,3}) \right]_{4,2}^{4,1}$$

$$\begin{bmatrix} x_m \\ y_m \\ z_m \end{bmatrix} + {}^f R_m \cdot \begin{bmatrix} r_m \cdot \cos(\beta_m) \\ r_m \cdot \sin(\beta_m) \\ 0 \end{bmatrix} = \left[ {}^f H_0^2 \cdot {}^0 H_1^2(q_{2,1}) \cdot {}^1 H_2^2(q_{2,2}) \cdot {}^2 H_3^2(q_{2,3}) \right]_{4,2}^{4,1} \quad (3)$$

$$\begin{bmatrix} x_m \\ y_m \\ z_m \end{bmatrix} + {}^f R_m \cdot \begin{bmatrix} r_m \cdot \cos(\beta_m) \\ -r_m \cdot \sin(\beta_m) \\ 0 \end{bmatrix} = \left[ {}^f H_0^3 \cdot {}^0 H_1^3(q_{3,1}) \cdot {}^1 H_2^3(q_{3,2}) \cdot {}^2 H_3^3(q_{3,3}) \right]_{4,2}^{4,1}$$

for points *A*, *B* and *C* respectively. ${}^f R_m$ is the rotation matrix of the mobile platform with respect to the fixed reference systems $\{O_f - X_f Y_f Z_f\}$. Subscript [4, 1..3] indicates that only the fourth column from rows 1 to 3 of the matrix is considered.

Equation (2) applied to the central limb,

$$\begin{aligned} x_m &= -\sin(q_{4,1}) \cdot q_{4,2} \\ y_m &= 0 \\ z_m &= \cos(q_{4,1}) \cdot q_{4,2} \end{aligned} \quad (4)$$

From these equations, we can easily obtain the active generalized coordinate $q_{4,2}$ and also the passive one, $q_{4,1}$. For the external limbs, a similar procedure can be followed in order to obtain explicit expressions for the generalized coordinates. For instance, in the case of limb 1, the active generalized coordinate and the first two passive coordinates can be obtained as follows,

$$\begin{aligned} a &= x_m^2 + z_m^2 + r^2 + r_m^2 + 2 \cdot r \cdot x_m + 2 \cdot r_m \cdot z_m \cdot sen(\theta) - 2 \cdot r_m \cdot x_m \cdot cos(\theta) \cdot cos(\psi) - \\ &\quad - 2 \cdot r \cdot r_m \cdot cos(\theta) \cdot cos(\psi) \\ b &= -x_m^2 - z_m^2 - r^2 - r_m^2 - 2 \cdot r \cdot x_m - 2 \cdot r_m \cdot z_m \cdot sen(\theta) + 2 \cdot r \cdot r_m \cdot cos(\theta) \cdot cos(\psi) + \\ &\quad + 2 \cdot r_m \cdot x_m \cdot cos(\theta) \cdot cos(\psi) + r_m^2 \cdot cos^2(\theta) \cdot sen^2(\psi) \end{aligned}$$

$$q_{1,3} = \sqrt{a} \quad (5)$$

$$q_{1,2} = atan2\left(\sqrt{-\frac{b}{a}}, \frac{r_m \cdot cos(\theta) \cdot sen(\psi)}{\sqrt{-b}}\right) \quad (6)$$

$$q_{1,1} = atan2\left(\frac{z_m + r_m \cdot cos(\theta)}{\sqrt{-b}}, \frac{r_m \cdot cos(\theta) \cdot sen(\psi)}{\sqrt{-b}}\right) \quad (7)$$

for the second stage, the remaining passive generalized coordinates, $q_{i,4}$, $q_{i,5}$, $q_{i,6}$, of the external limbs can be obtained from the equation as follows,

$$^f R_3(q_{i,1}, q_{i,2}, q_{i,3})\, ^3 R_6(q_{i,4}, q_{i,5}, q_{i,6}) = {}^f R_m(\phi, \theta, \psi) \quad i = 1 \text{L } 3 \quad (8)$$

*4 DoF parallel robot forward displacement*

For each of the robot's legs, the following vector closure equations can be established (see Figure (4)).

$$\vec{r}_{A_0A}(q_{1,1}, q_{1,2}, q_{1,3}) = \begin{bmatrix} x_m \\ 0 \\ z_m \end{bmatrix} + {}^fR_m(\theta, \psi) \cdot \begin{bmatrix} -r \\ 0 \\ 0 \end{bmatrix}$$

$$\vec{r}_{B_0B}(q_{2,1}, q_{2,2}, q_{2,3}) = \begin{bmatrix} x_m \\ 0 \\ z_m \end{bmatrix} + {}^fR_m(\theta, \psi) \cdot \begin{bmatrix} r \cdot \cos(\beta_{MD}) \\ r \cdot sen(\beta_{MD}) \\ 0 \end{bmatrix} \quad (9)$$

$$\vec{r}_{C_0C}(q_{3,1}, q_{3,2}, q_{3,3}) = \begin{bmatrix} x_m \\ 0 \\ z_m \end{bmatrix} + {}^fR_m(\theta, \psi) \cdot \begin{bmatrix} r \cdot \cos(\beta_{MI}) \\ -r \cdot sen(\beta_{MI}) \\ 0 \end{bmatrix}$$

$$\vec{r}_{O_fO_m}(q_{41}, q_{42}) = \begin{bmatrix} x_m \\ 0 \\ z_m \end{bmatrix}$$

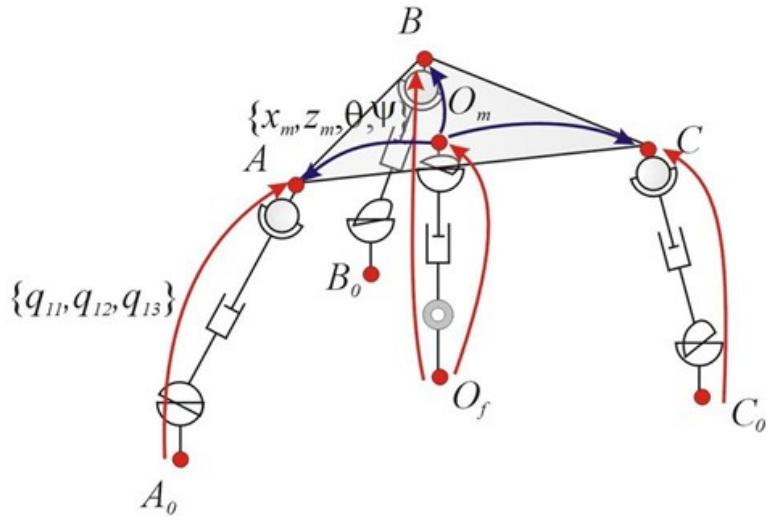

Figure 4. Close–loops of the manipulator.

From equation (9), a system of 11 non-trivial equations with 11 unknowns can be obtained. This system could be solved by means of the Newton-Raphson (N-R) numerical algorithm. However, in order to improve the calculation time and the convergence speed, the passive generalized coordinates will be eliminated from those equations, leading to a system of only four equations,

$$\Phi_1 = q_{1,3}^2 - r^2 - 2 \cdot r \cdot x_m + 2 \cdot r \cdot r_m \cdot cos(\psi) \cdot cos(\theta) - x_m^2 + \\ + 2 \cdot x_m \cdot r_m \cdot cos(\psi) \cdot cos(\theta) - z_m^2 - 2 \cdot z_m \cdot r_m \cdot sin(\theta) - r_m^2 = 0 \quad (10)$$

$$\begin{aligned}
\Phi_2 = {} & q_{2,3}^2 - r^2 + 2 \cdot r \cdot r_m \cdot \sin(\beta_{FD}) \cdot \cos(\beta_{MD}) \cdot \sin(\psi) \cdot \cos(\theta) + \\
& + 2 \cdot r \cdot r_m \cdot \sin(\beta_{FD}) \cdot \sin(\beta_{MD}) \cdot \cos(\psi) + 2 \cdot r \cdot x_m \cdot \cos(\beta_{FD}) + \\
& + 2 \cdot r \cdot r_m \cdot \cos(\beta_{FD}) \cdot \cos(\beta_{MD}) \cdot \cos(\psi) \cdot \cos(\theta) - \\
& - 2 \cdot r \cdot r_m \cdot \cos(\beta_{FD}) \cdot \sin(\beta_{MD}) \cdot \sin(\psi) - x_m^2 - \\
& - 2 \cdot x_m \cdot r_m \cdot \cos(\beta_{MD}) \cdot \cos(\psi) \cdot \cos(\theta) + 2 \cdot x_m \cdot r_m \cdot \sin(\beta_{MD}) \cdot \sin(\psi) - \\
& - r_m^2 - z_m^2 + 2 \cdot z_m \cdot r_m \cdot \cos(\beta_{MD}) \cdot \sin(\theta) = 0
\end{aligned} \quad (11)$$

$$\begin{aligned}
\Phi_3 = {} & q_{3,3}^2 - r^2 - r_m^2 + 2 \cdot r \cdot r_m \cdot \cos(\beta_{FI}) \cdot \cos(\beta_{MI}) \cdot \cos(\psi) \cdot \cos(\theta) + \\
& + 2 \cdot z_m \cdot r_m \cdot \cos(\beta_{MI}) \cdot \sin(\theta) + 2 \cdot r \cdot x_m \cdot \cos(\beta_{FI}) - 2 \cdot x_m \cdot r_m \cdot \sin(\beta_{MI}) \cdot \sin(\psi) - \\
& - 2 \cdot r \cdot r_m \cdot \sin(\beta_{FI}) \cdot \cos(\beta_{MI}) \cdot \sin(\psi) \cdot \cos(\theta) - z_m^2 + \\
& + 2 \cdot r \cdot r_m \cdot \cos(\beta_{FI}) \cdot \sin(\beta_{MI}) \cdot \sin(\psi) - x_m^2 - \\
& - 2 \cdot x_m \cdot r_m \cdot \cos(\beta_{MI}) \cdot \cos(\psi) \cdot \cos(\theta) + 2 \cdot r \cdot r_m \cdot \sin(\beta_{FI}) \cdot \sin(\beta_{MI}) \cdot \cos(\psi) = 0
\end{aligned} \quad (12)$$

$$\Phi_4 = q_{4,3}^2 - x_m^2 - z_m^2 = 0 \quad (13)$$

The N-R algorithm enables each set of active generalized coordinates of equations (10)-13) to be solved faster than the 11 coordinates system represented in equation (9). In order to avoid singular configurations, an asymmetrical array of the legs is proposed. Through a process of trial and error and considering the Range of Motion for LLR, the following values were selected for the geometric parameters of the PM: $r = 0.40$m, $r_m = 0.20$m, $\beta_{FD} = 50°$, $\beta_{FI} = 40°$, $\beta_{MD} = 40°$ and $\beta_{MI} = 30°$.

**Mechatronic Manipulator Development**

Four DC motors equipped with power amplifiers have been used to actuate the 4 DoF PM. The actuators are *Maxon RE40 Graphite Brushes 150W* motors. These high-quality motors are fitted with powerful permanent magnets and an ironless rotor, as well as being compact, powerful, low-inertia 150 Watt motors. The performance specifications of these *Maxon's* motors are 24V nominal voltage, 6940rpm nominal speed, 6A max. continuous current and 2420mNm stall torque. The characteristics of the motor matches the actuation requirement.

These actuators are equipped with encoder sensors and brakes. The encoder sensor is the *ENC DEDL 9149* system which is a digital incremental encoder with 500 pulses per revolution, 3 channels and 100 kHz max. operating frequency. The brake system is the *Brake AB 28 system*, which is a 24 V, 0.4 Nm permanent-magnet, single-face brake for DC motors that prevents rotation of the shaft at standstill or when the motor power is turned off.

*Hardware control architecture*

An industrial PC and a power amplifier stage have been used to implement the control architecture for this PM (see Figure 5). The PC is based on a high performance 4U Rackmount industrial system with seven PCI slots and seven ISA slots. It has a 3.10GHz Intel ® CORE i7 processor and 4 GB DDR3 1333 MHz. SDRAM. The industrial PC is equipped with two *Advantech$^{TM}$* data acquisition cards: *PCI-1720* and *PCI-1784*. The *PCI-1720* card has been used to supply the control actions for each parallel robot actuator, providing four 12-bit isolated digital-to-analog outputs for the Universal PCI 2.2 bus. The card has multiple output ranges (0~5V, 0~10V, ±5V, ±10V), a programmable software and 2500 VDC isolation protection between the outputs and the PCI bus. The *PCI-1784* card is a 4-axis quadrature encoder and counter add-on card for the PCI bus. The card includes four 32-bit quadruple AB phase encoder counters, an onboard 8-bit timer with a wide range time-based selector and it is optically isolated up to 2500V.

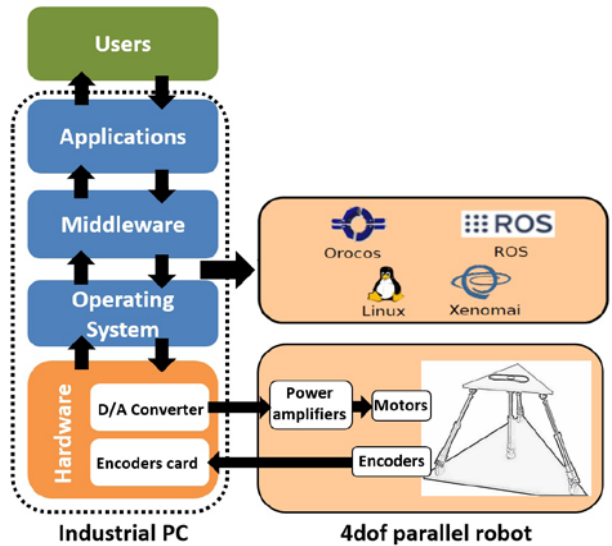

Figure 5. Manipulator control architecture.

An amplifier unit has been developed to control the Maxon's motors. It consists of three stages: an analog to Pulse Width Modulation (PWM) stage, an H-bridge gate driver, and a FETs stage (see Figure 6). The first stage transforms the analog voltage supplied by the PC control into a PWM. The analog to PWM stage is based on an *LTC6992* silicon oscillator (*TimerBlox*®). The output frequency is determined by a single resistor that programs the *LTC69920's* internal master oscillator frequency.

The PWM signal and the movement sense (provided by a digital output from the PC control) are supplied to the H-bridge gate driver, which is based on the *DRV8701* device from *Texas Instruments*® with a brushed DC motor full-bridge driver that uses four external N-channel MOSFETs targeted to drive a 12V to 24V bidirectional brushed DC motor.

Finally, the power amplifier unit has four MOSFETs in a full H-bridge configuration.

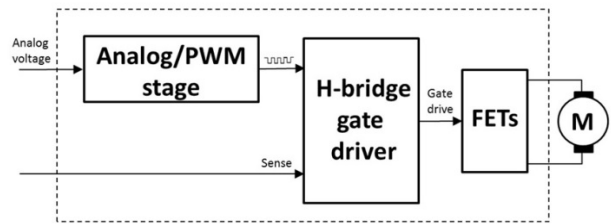

Figure 6. Power amplifier stage.

*Software control architecture*

The software architecture represents one of the critical aspects when implementing a new robot system. In recent years, there has been an increase in component-based software development due to the following advantages:

- Modular design and structure.
- Fully reusable code and modules.
- Reconfigurable modules.
- Distributed execution of the modules, improving total execution time

Since different control schemes share common parts, the modular design consists of developing each part as a module, thus ending up with several modules. The developer then uses these modules to implement different controllers as if building a puzzle. The developer can configure (making connections between modules) and run the control scheme by inserting the necessary modules. Note that, although developing the modules can be a complicated task at first, the component-based software makes the programmer's job easier in the long run because if a module works correctly in one particular scheme, it will certainly work as well in another control scheme. In addition to the advantages discussed above, this approach minimizes the chance of programming errors in the implementation of any of the modules.

The control architecture (see Figure 5) uses an industrial PC with Linux Ubuntu 12.04 operating system. The rehabilitation therapy requires that the control scheme is able to be run real-time, which can be obtained by means of the real-time kernel patch Xenomai. The proposed control architecture presents two main advantages: 1) The architecture allows us to eventually implement and programme any required control algorithm, as well as allowing us to use external sensors, such as artificial vision, cameras, force sensors and accelerometers by only plugging the appropriate module. 2) The control architecture is low-cost because the programming platform was built with free software tools. Including the cost of an industrial PC equipped with industrial data acquisition cards, the cost remains below $2000.

The robot control algorithms are developed by taking advantage of the middleware *Open Robot Control Software* (*Orocos* (Bruyninckx, 2002)) and *Robot Operating System* (*ROS* (Garage, 2009)). Nowadays, Orocos represents one of the best real-time motion control frameworks available, but it does have certain constraints when trying to achieve something other than control itself. One of the solutions is *ROS*, which was designed as a conglomeration of various tools organized in packages. Each package (or "stack") may contain libraries, executables or scripts and a manifest which defines the dependencies on other packages and meta information about the package itself. A *ROS* package called *rtt ros integration* allows *Orocos* components to connect to the *ROS* network making both middleware fully compatible.

Concisely, ROS provides many tools and functionalities which are useful when developing robotic applications, while Orocos provides a solid core for real-time control. Both types of software complement each other and widen the range of applications they can offer as standalone platforms.

**Control of the 4 DoF PM**

The control of the PM can be developed through different control strategies. For instance, model-based controllers which compensate for the nonlinearities of the robot (such as inertial, gravitational and Coriolis terms) by adding these forces to the control action. These kinds of controllers have two main problems. First, they are more difficult to program and have greater computational complexity. Secondly, model-based controllers require the model dynamic parameters, and therefore a parameter identification process is needed (Díaz-Rodríguezet al., 2010).

In this paper, a passivity-based controller has been developed to control the novel 4 DoF PM. The passivity-based approach solves the control problem by taking advantage of the passivity property of the robot system's physical structure by reshaping the natural energy of the system in such a way that the tracking control objective is achieved (Ortega and Spong, 1989).

The control algorithm is based on the work of (Ortega et al., 2013). The control law obeys the following equation:

$$\tau_c = -K_p \cdot e - K_d \cdot v - Ki \cdot \int_0^t (e+v)dt \qquad (14)$$

where $K_p$, $K_d$ and $K_i$ are positive definite diagonal matrices. The controller which offers significant system performance and robustness properties is a PID. This controller has proportional, derivative and integral components. The first calculates the error between the active generalized coordinates and their references ($e=q-q_d$). The active coordinates values of the linear actuators are measured using the encoder card. The derivative component depends on the velocity of the joints, and because the proposed robot does not provide velocity sensors, the velocity measurement for this controller has been replaced by approximate differentiation:

$$v = diag\left\{\frac{b_i s}{s + a_i}\right\} \cdot q \tag{15}$$

with $a_i > 0$ and $b_i > 0$. Finally, the controller provides an integral component which is introduced in the control law as a standard practical remedy to compensate for the robot gravity term.

This control algorithm has been developed in the open control architecture using the programmed OROCOS/ROS modules (see Figure 7). The *Cartesian Reference* module calculates the movement references in the Cartesian plane, and the *Inveser Kinematics* module obtains the references for the four active joints of the robot coordinates ($q_{13\_ref}$, $q_{23\_ref}$, $q_{33\_ref}$ and $q_{24\_ref}$). The robot coordinates are obtained by the *Encoders Card PCL-1784* module. The *Velocity Estimation* module provides the robot velocity estimation following the equation (15). The *PID Controller* module calculates the control action depending on the proportional, derivative and integral terms, which it then provides to the actuator module which is in turn responsible for carrying out digital-analog conversions through the Advantech PCI-1720 card.

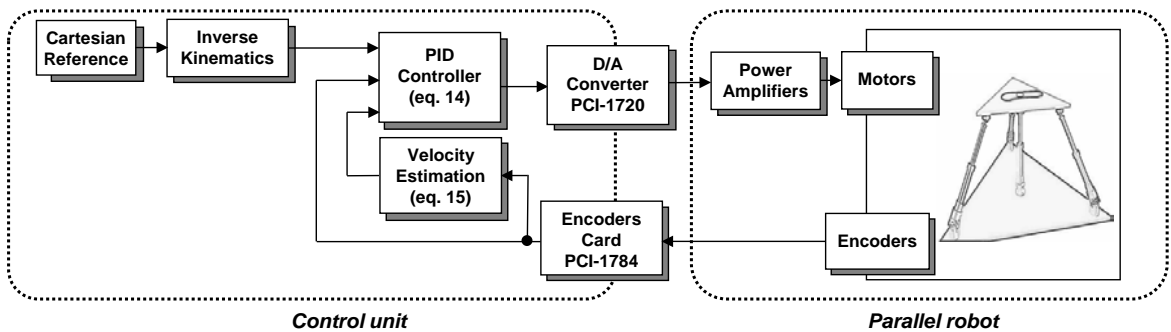

Figure 7. Passivity-based controller implementation in the open control architecture.

In order to validate the robot design and control architecture, several trajectories have been tested. Due to the space limit, only two of them are included in the paper. Figure 8 shows the references for a first execution. In this case, the reference for the Z coordinate and the yaw orientation are based on a sinusoidal motion. The references for the *X* coordinate and the pitch orientation remain motionless.

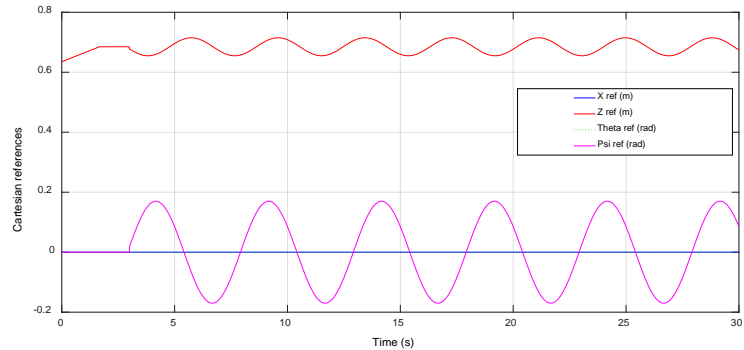

Figure 8. Cartesian reference for the 1st movement of the robot.

Figure 9 presents the response of the four active coordinates of the parallel manipulator, according to the Cartesian references proposed above. The first column of this figure shows the joint references (obtained by the inverse kinematics of the robot using equation (5-8)) and the robot joint positions. The second column shows the position error. As we can clearly see, the manipulator follows the required trajectory with very small mean errors (see Table 3). In addition, the phase offset has been calculated according to Ramsay and Silverman, 1997, where the value is very low (41.5 ±7.0 ms) and shows that the controller presents a very fast response which, in all cases, is lower than the human time reaction (more than 150.0 ms).

Figure 10 shows (in blue) the references for a second execution. In this case, the reference is an elliptic motion in the *X-Z* plane. Before the periodic motion, the centre of the mobile platform follows a linear motion path from the origin (0, 0, 0.635) to the position (0.05, 0, 0.69), and then a second movement on the *Z* axis to the point (0.05, 0. 0.75). The actual robot response is represented in black.

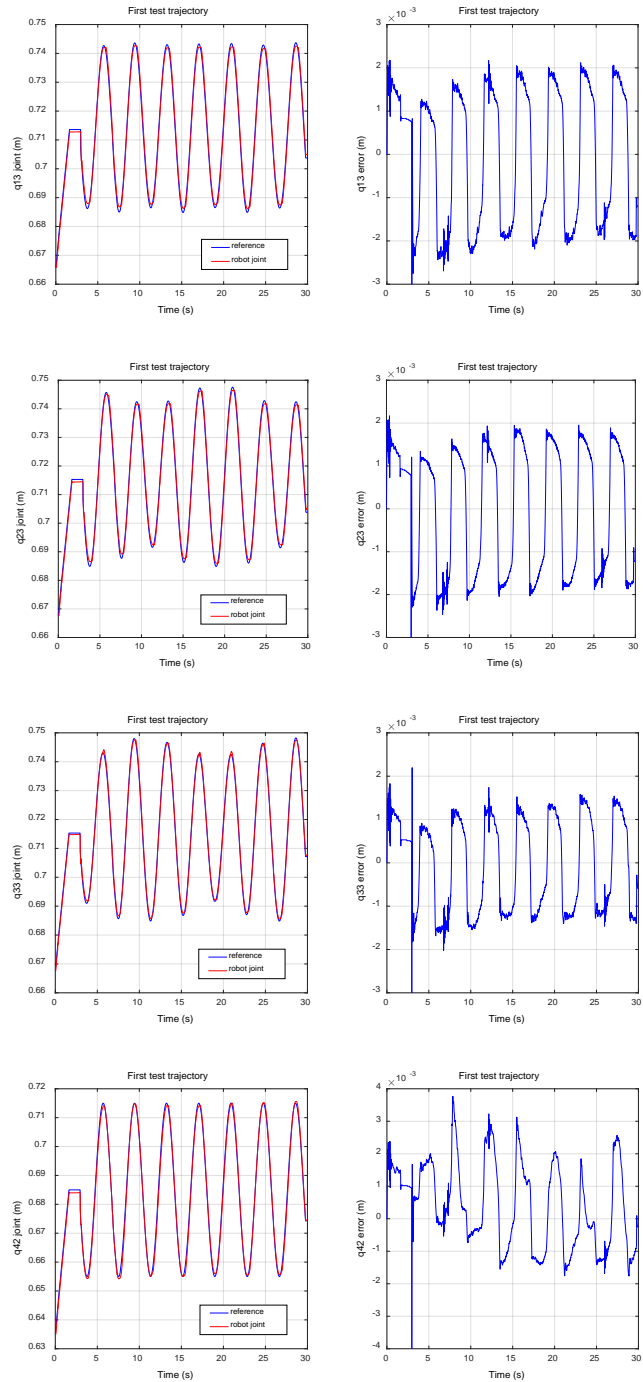

Figure 9. Active robot coordinates and position errors.

Figure 10 shows (in blue) the references for a second execution. In this case, the reference is an elliptic motion in the *X-Z* plane. Before the periodic motion, the centre of the mobile platform follows a linear motion path from the origin (0, 0, 0.635) to the position (0.05, 0, 0.69), and then a second movement on the *Z* axis to the point (0.05, 0. 0.75). The actual robot response is represented in black.

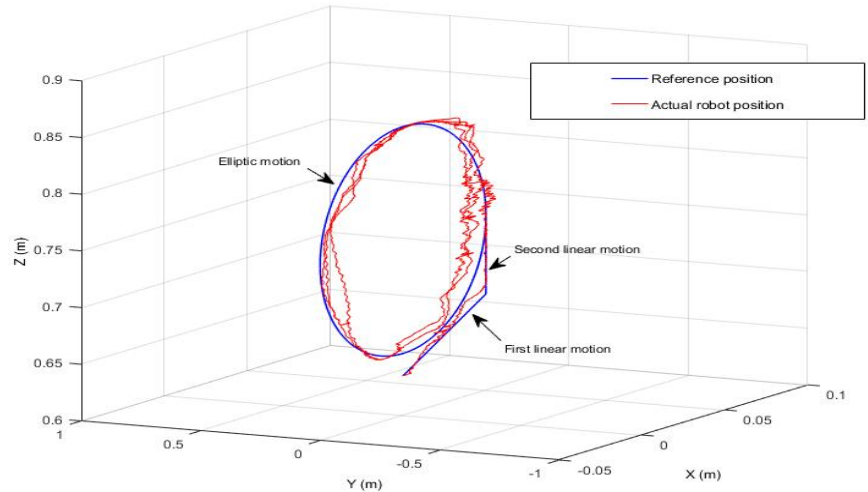

Figure 10. Cartesian reference and actual robot position for the second trajectory

Figure 11 presents the response of the four active coordinates of the parallel manipulator for the second trajectory test. This figure shows the joint references and the robot joint positions as well as the control action applied. As in the first test trajectory, the manipulator response accomplishes the task.

Table 3 shows the difference in the reference value and the actual PM active joints for the two movements presented in Figures 8 and 10. This difference is described by the mean error value. As shown in this table, the control algorithm implemented gives a very low error, which means that the system achieves the specified reference without any problems.

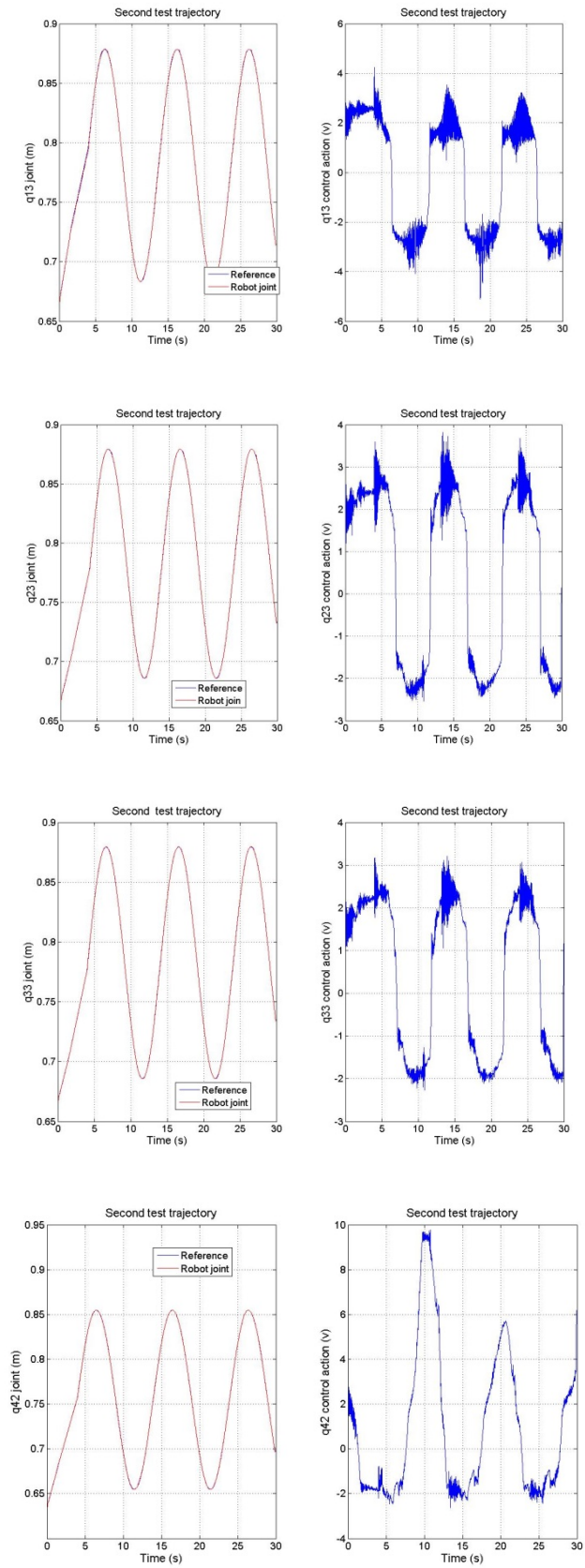

Figure 11. Active robot coordinates and control actions.

**Table 3** Mean errors (m)

|  | joint 13 | joint 23 | joint 33 | joint 42 |
|---|---|---|---|---|
| 1$^{st}$ movement | -4.570e-5 | -2.0528e-5 | -2.8629e-5 | 4.3646e-5 |
| 2$^{nd}$ movement | -8.961e-5 | -4.6443e-5 | -1.082e-4 | 3.343e-4 |

**Conclusions**

This paper has shown the development of a novel low-cost 4 DoF parallel manipulator. The PM design was based on the need to develop a lower limb rehabilitation system. The developed manipulator allows us to apply combinations of normal or tangential efforts in the leg joints, as well as torque acting on the knee which, to the authors' best knowledge, enhances previous designs. We have fully developed the mechatronic design, mechanical structure, electromechanical actuators and control system. In this regard, we have developed a new open control architecture for the manipulator control. The control hardware is based on an industrial PC equipped with industrial data acquisition cards which read the manipulator joint positions and provide the actuator with the control actions through digital-to-analog converters. The software architecture is based on free and open source software: OROCOS and ROS middleware. The proposed control architecture has two main advantages. First, the open control architecture allows us to eventually implement and program any required control algorithm by only plugging in the appropriate module. Secondly, the price of this control system remains below 2000$.

The control of a rehabilitation task should be carried out in task space. Thus, we presented the direct and inverse kinematic equations for the PM which are programmed into the control unit as a part of the passivity-based control scheme. The control algorithm is a point-to-point controller that uses an estimation of the robot's velocity and an integral action to cancel the gravitational term of the robot. Different results demonstrating the tracking accuracy of the proposed controller have been included, showing an accurate response in terms of position error. Finally, we have presented a step by step approach in a didactic way, which can serve as an interesting reference for others to follow on the mechatronics design of PMs.

**Acknowledgements**

The authors wish to thank the Plan Nacional de I+D, Comisión Interministerial de Ciencia y Tecnología (FEDER-CICYT) for the partial funding of this study under project DPI2013-44227-R. We also want to thank the Fondo Nacional de Ciencia, Tecnología e Innovación (FONACIT-Venezuela) for it financial support under the project Nº 2013002165.